\documentclass[10pt,twocolumn,letterpaper]{article}

 \usepackage{cvpr}              

\usepackage{tabularx}
\usepackage{stfloats}
\definecolor{cvprblue}{rgb}{0.21,0.49,0.74}
\usepackage[pagebackref,breaklinks,colorlinks,allcolors=cvprblue]{hyperref}
\usepackage{graphicx}
\usepackage{tabularx}
\usepackage{array}

\def\paperID{*****} 
\def\confName{CVPR}
\def\confYear{2026}

\title{Attention Is not Everything: Efficient Alternatives for Vision}

\author{
Nur Mohammad Kazi\\
Ahsanullah University of\\
Science and Technology\\
Dhaka, Bangladesh\\
{\tt\small nurmohammadkazi2002@gmail.com}
\and
Ibteshum Khaled\\
Ahsanullah University of\\
Science and Technology\\
Dhaka, Bangladesh\\
{\tt\small ibteshum.khaled123@gmail.com}
\and
Md. Luthful Hasan Galib\\
Ahsanullah University of\\
Science and Technology\\
Dhaka, Bangladesh\\
{\tt\small omi673galib@gmail.com}
\and
Ali Faruk Shihab\\
Ahsanullah University of\\
Science and Technology\\
Dhaka, Bangladesh\\
{\tt\small faruk.aka.shihab@gmail.com}
\and
Md. Rakibul Islam\\
Ahsanullah University of\\
Science and Technology\\
Dhaka, Bangladesh\\
{\tt\small rakib.cse@aust.edu}
}

\begin{document}
\maketitle
\begin{abstract}
Recently computer vision has seen advancements mainly thanks to Transformer-based models. However many non-Transformer methods are still doing well being a direct competition of Transformer-based models. This review tries to present a comprehensive taxonomy of such methods and organize these methods into categories like convolution-based models, MLP-based models, state-space-based and more. These methods are looked at in terms of how efficient they are, how well they scale, how easy they are to understand and how robust they are. A total of 40 papers were chosen for this study. The goal is to give a view of non-Transformer methods and find out what challenges and opportunities exist for future computer vision research.

\end{abstract}
\vspace{2mm}
\section{Introduction}
\label{sec:intro}

Computer vision is an ever changing field with different ways of thinking and development of new models. At first people used a lot of manual help and algorithms and then deep learning came along. Convolutional neural networks became popular because they had better understanding of relationships in a space than previous methods and relied less on manual input. Recently people have been using models that are good at understanding how things change over time and models that can handle continuous signals, like sound or video. They are also using methods to optimize things and frameworks that use probability to estimate how sure we are about something. People are also combining computer vision methods with deep learning to make models that are easier to understand and can work better in different situations.

\begin{figure}[h]
  \centering
  \includegraphics[width=0.47\textwidth]{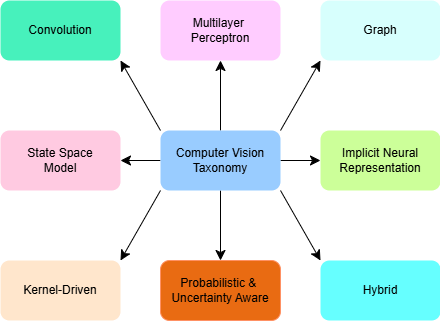}
  \caption{Computer Vision Taxonomy}
  \label{fig:computer_vision_taxonomy}
  
\end{figure}

The introduction of Transformer architectures has significantly influenced modern machine learning fields, specifically computer vision. By enabling global context modeling through self-attention mechanisms, Transformers achieved impressive results that reshaped the research field. Starting with the work “Attention Is All You Need” \cite{r1}, Transformers were adapted to vision tasks through models such as the Vision Transformer (ViT) \cite{r2}, which demonstrated that pure attention-based architectures can achieve competitive performance when trained on large-scale datasets. Following works, including Data-efficient Image Transformers (DeiT) \cite{r3} and the Swin Transformer \cite{r4}, further improved performance establishing Transformers as a dominant paradigm in the field. However methods that use Transformers have some problems. They need a lot of data to work well. They use a lot of computational power and memory. They do not have built-in functionality that helps when there is not much data available. Non-Transformer models complements them as they are capable of mitigating stated limitations.
Non-transformer approaches remain highly relevant and continue to evolve as competitive alternatives. It can be highly practical when transformer based models are not compatible with the given task environment. This paper presents a comprehensive taxonomy of such methods, including convolution-based models, MLP-based models, graph-based models, state-space and structured models, implicit neural representations, energy-based and kernel-driven methods, probabilistic frameworks, and classical CV–deep hybrid approaches. This paper presents a comprehensive taxonomy given in figure \ref{fig:computer_vision_taxonomy}.

By analyzing these paradigms, we aim to provide a structured perspective on the landscape of computer vision. These papers were chosen in order to guarantee both relevance and representation across the eight computer vision taxonomies identified in this literature review. In all, 40 papers have been selected, ranging from 2023 to 2025, by using an approach that includes representing the most recent achievements, thus making it possible to cover both classical approaches and advanced techniques. This selection process has been done considering several criteria, including the methodological importance, relevance, and citations of each paper to its corresponding taxonomy.

\section{Convolution-Based Approach}
\label{sec:approach1}

The principal features that have made CNNs the longstanding paradigm of the computer vision are their exploitation of spatial locality and weight sharing which reduces the hierarchical feature acquired in an efficient manner. Even though Vision Transformers have been shown to be better in terms of their global self-attention mechanisms, intensive execution limitations placed by their quadratic computational complexity restrict scalability in a great way. Current CNN studies have hence adapted bigger kernels, deformable operators and transformer-liked block structures to close this performance disparity with maintaining the intrinsically performance efficient convolutional operations.

\begin{table*}[t]
\centering
\small
\caption{Summary of CNN-based Architectures}
\label{tab:vision_comparison}
\begin{tabularx}{\textwidth}{l p{2cm} p{2.2cm} X X X}
\toprule
\textbf{Paper} & \textbf{Model} & \textbf{Vision Task} & \textbf{Contribution} & \textbf{Strength} & \textbf{Limitation} \\ \midrule

Liu et al. (2023) \cite{r5} & ConvNeXt V2 & Classification, Detection, Segmentation & FCMAE + sparse convolutions; GRN layer to prevent feature collapse & No extra FLOPs from GRN; scalable from Atto to Huge & Without GRN, masked training still underperforms; ViT edges out at huge scale \\ \addlinespace

Wang et al. (2023) \cite{r6} & InternImage & Classification, Detection, Segmentation & DCNv3 core operator; first CNN scaled to 1B+ parameters & Adaptive receptive field; efficient sparse sampling; inherits CNN inductive bias & Scaling requires careful tuning of four hyper-parameters under strict stacking rules \\ \addlinespace

Ding et al. (2024) \cite{r9} & UniRepLKNet & Classification, Detection, Segmentation, Multi-modal & Four structural guidelines for large-kernel ConvNets; Dilated Reparam Block; universal multi-modal backbone & State-of-the-art across vision and non-vision modalities; higher shape bias than ViTs & Parallel dilated branches increase training cost; generative/VLM tasks unexplored \\ \addlinespace

Yu et al. (2024) \cite{r11} & InceptionNeXt & Classification, Segmentation & Decompose large-kernel depthwise conv into four parallel branches (3x3, 1x11, 11x1, identity) & Linear parameter/FLOP cost; 1.6x higher throughput than ConvNeXt-T & Simplified block design limits self-attention integration \\ \addlinespace

Wang et al. (2024) \cite{r12} & RepViT & Classification, Detection, Segmentation & Modernize lightweight CNN via MetaFormer block structure and structural re-parameterization & First lightweight model $>$80\% accuracy at 1ms latency; $\sim$10x faster than MobileSAM & Avoids large kernels (7x7+) due to mobile compiler limits \\ 
\bottomrule
\end{tabularx}
\end{table*}

\subsection{State-of-the-Art Convolution-Based Methods}

At Liu et al. (2023) ConvNeXt V2 \cite{r5} was proposed, to deal with the incompatibility between canonical ConvNets and masked Autoencoders (MAE). Traditional sliding-window convolutions create blank spaces in the image, which results into a redundant, or nonfunctional neuron, in situations where 60 \% of an image is masked, which is known as feature collapse. The proposed Fully Convolutional Masked AutoEncoder (FCMAE) addresses this problem by the use of sparse convolutions that run only over visible pixels. An important output of FCMAE is its Global Response Normalization (GRN) layer that drives channels to compete against each other and hence yields channels to learn a different representation without extra overhead.
With the help of a lightweight decoder, the masked areas are then reassembled. The available empirical evidence confirms that with Top-1 accuracy is 76.7\% on the Atto variant vs 88.9\% on the huge one thus proving that the pure ConvNets remain competitive when in combination with appropriate self-supervision-training plots.  
 
The first CNN-based foundation model with more than a billion parameters, introduced by Wang and colleagues (2023) is known as InternImage \cite{r6}. The traditional CNNs are constrained by fixed receptive field, and fixed weight that limits them to learn the long-range dependencies that provide large scale-effectiveness in Vision Transformers. To get results at both ends, InternImage uses something called Deformable Convolution version 3. This is where a small 3 by 3 kernel is kept and some other things, like sampling offsets and modulation scalars are changed in a flexible way based on the InternImage input content. InternImage does this to make sure things work well with the InternImage. The features of different representational subspaces are captured at once by a multi-group spatial aggregation mechanism. Combined with operation-inspired block designs and simulated stacking rules, InternImage-H achieves 89.6 About-1 accuracy on ImageNet\cite{r7}, 65.4 box AP on 6 CoC coC test-dev, and 62.9 IoU on 80 Automated Detection and localization Around 20 chest, called ADE20K\cite{r8}.  
 
Ding and others (2024) came up with UniRepLKNet\cite{r9}, which is based on principles that are explicitly designed to be applied to deep large-kernel ConvNets instead of using the principles of a transformer or even traditional CNNs. It introduced four rules of structure: (1) use of SE blocks to scale depth by the use of a network with parallel dilated small-kernel branches embedded in the network at training, then in inference the small kernels are interconnected into one large-kernel layer; (2) the use of a Dilated Reparameterization Block, which compiles parallel dilated small-kernel branches in the network during training and then combines them into one large-kernel layer during inference; (3) layout of large kernels on middle and high-level layers; and (4) the replacement of small to support multimodal capability, audio, video and point-cloud information are converted to multimodal format in a modality-specific preprocessing step and embedded in 3D maps, which are processed using a common backbone. UniRepLKNet has 88.0\% ImageNet accuracy, 55.6\% mIoU on ADE20K\cite{r8}, and 98.5\% accuracy on Speech Commands V2\cite{r10}.

Yu and co-authors (2024) found that the ConvNeXt style of building things uses really big depthwise convolutions. These convolutions are very expensive when it comes to getting information from memory. This makes them really slow when they do not have to do a lot of math with decimal numbers. The ConvNeXt style is affected by this problem. InceptionNeXt \cite{r11} is a solution that is able to break up a large-kernel operation into four parallel streams along the channel dimension: a 311 square-kernel, a 11 vertical band, a 1 horizontal band, and an identity transform. The resulting outputs are simply concatenated resulting in an approximation of the representational power of a large kernel at particular fraction of the cost since channel counts and kernel size grow linearly. InceptionNeXt -T is equivalent to ConvNeXt -T in ImageNet, but has 1.6 times better training throughput.  

Wang et al. presented RepViT\cite{r12}, that modernizes the lightweight MobileNetV3-L with systematic absorption of Vision Transformer architectural understanding. The redesign isolates the token mixer and channel mixer into a MetaFormer style block architecture and implements structural re-parameterization to combine multi-branch training time representations into a layer which can be run in one efficiency at test time. Other design options are convolutional stem, finer downsampling layers, a simpler classifier head and cross-block Squeeze-and-Excitation. RepViT -M1.0 is the first lightweight model to achieve top-1 accuracy above 80\% on ImageNet even at the low latency of 1.0ms on an iPhone 12, and RepViT -SAM is nearly ten times faster than MobileSAM.

\begin{table*}[b]
\centering
\small
\caption{Summary of MLP-based Vision Models}
\label{tab:mlp_vision_comparison}
\begin{tabularx}{\textwidth}{l p{2cm} p{2.2cm} X X X}
\toprule
\textbf{Paper} & \textbf{Model} & \textbf{Vision Task} & \textbf{Contribution} & \textbf{Strength} & \textbf{Limitation} \\ \midrule

Liao et al. (2023) \cite{r13} & ALOFT & Domain Generalization & Dynamic low-frequency spectrum transform to suppress texture bias and simulate domain shifts & State-of-the-art DG at $\sim$14M parameters; effective non-parametric frequency augmentation & Assumes Gaussian spectrum distribution; sensitive to perturbation hyperparameters \\ \addlinespace

Huang et al. (2023) \cite{r15} & Strip-MLP & Image Classification & Cross-strip token interaction; CGSMM for deep-layer interaction independent of spatial resolution & Solves token interaction dilemma; +2.44\% on Caltech-101, +2.16\% on CIFAR-100 & Complex patch splitting and channel permuting add structural overhead \\ \addlinespace

Yang et al. (2022) \cite{r17} & Dynamic MLP & Fine-Grained Classification & Bilateral-path architecture with metadata-conditioned dynamic projection weights & High-dimensional image-metadata interaction; end-to-end trainable; competitive inference speed & Degrades without metadata; slight parameter and FLOP increase over image-only baseline \\ \addlinespace

Meng et al. (2024) \cite{r19} & CorrMLP & Medical Image Registration & First MLP-based coarse-to-fine registration; CMW-MLP block with 3D correlation and multi-window branches & Full-resolution processing; outperforms CNNs and Transformers; $<$0.5s inference on GPU & Large 3D models still memory-demanding; adapting 3D variants non-trivial \\ \addlinespace

Zheng et al. (2024) \cite{r20} & RivuletMLP & Video Quality Enhancement & DDA for temporal alignment, SFF for global dependencies via permutation, BSC for local refinement & 47.5\% faster than CPGA; stable quality across frames; reduces texture flickering & Slightly higher parameter count than lightweight CNN-only models \\ 
\bottomrule
\end{tabularx}
\end{table*}

\section{MLP-Based Approach}
\label{sec:approach2}
Multi-Layer Perceptron (MLP)-based vision architectures showed that competitive visual representations can be learned without convolutions or self-attention. Limitations of this approach is mitigated by incorporating frequency-domain operations, cross-strip interactions, dynamic weight generation, and domain-shift augmentation. 

\subsection{State-of-the-Art MLP-Based Methods}
Liao et al. (2023) proposed ALOFT \cite{r13} to address a core limitation of CNN-based domain generalization: local convolutions cause models to over-rely on texture, degrading performance on unseen target domains. ALOFT builds on a Global Filter Network baseline and introduces a Dynamic Low-Frequency Spectrum Transform module that simulates domain shifts in the Fourier domain. Low-frequency components carry the most domain-specific texture information; ALOFT models their cross-sample distribution and randomly resamples new spectra to replace the originals, compelling the network to rely on global structural content. Two variants are offered: element-level (ALOFT-E) and statistics-level (ALOFT-S). On the PACS\cite{r14} benchmark, ALOFT-E achieves 91.58\% accuracy, outperforming the strongest CNN baseline MVDG by 5.02 percentage points with a comparable ~14M parameter budget.

Huang et al. (2023) identified the token interaction dilemma in MLP vision models \cite{r15}: as feature maps are spatially down-sampled in deeper layers, the number of available tokens shrinks, severely weakening expressiveness. Strip-MLP introduces a Strip MLP layer that enables cross-strip interaction, where tokens in a given row or column contribute to the aggregation of adjacent strips. To decouple interaction capacity from spatial resolution, a Cascade Group Strip Mixing Module (CGSMM) splits data along the channel dimension and performs within-patch and cross-patch mixing. A Local Strip Mixing Module (LSMM) strengthens short-range interactions using a compact Strip MLP unit. Strip-MLP achieves +2.44\% on Caltech-101\cite{r107} and +2.16\% on CIFAR-100\cite{r16} over existing MLP baselines while matching state-of-the-art methods on ImageNet\cite{r7}.

Yang et al. (2022) addressed fine-grained visual classification, where species share nearly identical visual appearances \cite{r17}. The key observation is that location and date metadata strongly constrain plausible species identity, yet standard fusion methods such as concatenation operate in a single shared dimension and are insufficient for high-dimensional interaction. Dynamic MLP introduces a bilateral-path architecture: one path processes the image through a standard CNN while the other encodes metadata via a residual MLP. Crucially, the metadata path dynamically generates projection weights that transform the image representation to reflect geographic and seasonal priors. The framework achieves 91.39\% top-1 accuracy on iNaturalist 2021\cite{r18} and placed 3rd at the global iNaturalist challenge at CVPR 2021.

Meng et al. (2024) presented CorrMLP\cite{r19}, the first MLP-based framework for deformable medical image registration. Precise pixel-to-pixel alignment across full-resolution 3D scans requires both global context and fine-grained local detail, yet Transformers are too memory-intensive at full resolution while CNNs lack the inductive bias for local spatial correlation. CorrMLP introduces a Correlation-aware Multi-Window MLP (CMW-MLP) block comprising a 3D correlation layer and parallel MLP branches with window sizes of 3×3×3, 5×5×5, and 7×7×7 to capture a range of motion scales. A coarse-to-fine architecture refines registration progressively using image-level and step-level correlations. Training is unsupervised via a smoothness-penalized image dissimilarity loss. CorrMLP outperforms ten competing methods on brain and cardiac MRI benchmarks with inference under 0.5 seconds on a GPU.

Zheng et al. (2024) proposed RivuletMLP\cite{r20} to restore visual quality in videos degraded by HEVC lossy compression. CNNs suffer from limited receptive fields while Transformers are too expensive for real-time use. RivuletMLP relies on three purpose-built modules: Dynamically Guided Deformable Alignment adaptively aligns frames without accumulating errors; Spatiotemporal Feature Flow (SFF) uses a feature permutation strategy to relocate distant features into the central perception range, establishing global dependencies efficiently; and Benign Selection Compensation (BSC) corrects motion discontinuities via local-region refinement. On the MFQEv2.0\cite{r21} benchmark, RivuletMLP achieves an average 0.11 dB PSNR gain over top models at 10.0 FPS on 720p video, running 47.5\% faster than the transformer-based CPGA baseline.

\section{Graph-Based Approach}
\label{sec:approach3}
Graph is another term for a network. Graphs contain objects which are called Nodes or Vertices.These nodes/vertices are connected together. The connections between the vertices are called edges. Graph Neural Networks(GNN) are a class of deep learning models that process data structured as graphs. This approach helps to capture complex relationships and dependencies between nodes. Although it was not originally designed for computer vision tasks, ViG\cite{r22} and other graph based approaches show great promise in complex vision tasks.
\begin{table*}[t]
\centering
\small
\caption{Summary of Graph-based Vision Models}
\label{tab:graph_vision_comparison}
\begin{tabularx}{\textwidth}{l p{2cm} p{2.2cm} X X X}
\toprule
\textbf{Paper} & \textbf{Model} & \textbf{Vision Task} & \textbf{Contribution} & \textbf{Strength} & \textbf{Limitation} \\ \midrule

Han et al. (2023) \cite{r23} & HGNN & Classification and Object Detection & Higher order relationship modeling via hypergraphs & Reduced redundancy in feature representation & Manual hypergraph size determination \\ \addlinespace

Munir et al. (2024) \cite{r25} & GreedyViG & Classification, Detection, Segmentation & More efficient graph construction for Vision GNNs & Adaptive and task-specific architecture & Use of estimation reduces trustness/reliability \\ \addlinespace

Tian et al. (2024) \cite{r26} & IPG & Image Reconstruction & SR effective aggregation techniques & Computationally efficient and cheap & Low inference speed compared to simpler baselines \\ \addlinespace

Zhang et al. (2023) \cite{r29} & AF-GCN & Point Cloud Segmentation & New representation specifically for point cloud geometry & Suppresses noise and addresses isotropic kernels limitations & Highly dependent on the initial feature learning phase \\ 
\bottomrule
\end{tabularx}
\end{table*}

\subsection{State-of-the-Art Graph-Based Methods}
Instead of using spatial grid or sequence tokens, ViG treats image patches like a node in a graph. Images are divided into such nodes and they are projected into a feature vector through a linear embedding layer. The edges are built using k-nearest neighbors, connecting each node to its k most similar counterparts. A grapher module is used so node embedding reflects both local and global relationships. Feed-Forward Neural Network(FFN) are used to reduce number of nodes while increasing feature dimensionality. Finally, the node features are aggregated and used as needed for vision tasks.

As the previous approach confines itself to simple pairwise relationships, the researchers enhanced ViG. Han et al. (2023) built ViHGNN\cite{r23} by harnessing the power of the hypergraph to encompass more intricate inter-patch associations. ViHGNN uses Fuzzy C-Means clustering to group these nodes into hyperedges. Because Fuzzy C-Means allows for overlapping clusters, a single node can belong to multiple hyperedges, enabling the model to represent complex inter-object relationships. The surplus edges are removed which is a major state-of-the-art limitation. ViHGNN hyperedges can link any number of nodes capturing high-order relationships while reducing computational expenses and unwarranted memory. The model's effectiveness is rooted in a fundamental dependency: generating a robust hypergraph depends on high-quality patch embeddings, but those embeddings cannot be effectively generated without an existing hypergraph structure. The model achieves state-of-the-art results, including a 83.9\% top-1 accuracy on ImageNet and a 43.1\% Average Precision (AP) on COCO object detection\cite{r24}.

Munir et al. (2024) introduced Dynamic Axial Graph Construction, which mitigates a major bottleneck in ViGs that is the inefficient k-nearest neighbor (KNN) operation used for graph construction. GreedyViG\cite{r25}, which uses DAGC, estimates the mean and standard deviation instead of calculating the true similarity between every node.  It achieves this efficiently by splitting an image into quadrants and comparing the original image to one where the quadrants have been flipped across the diagonal. To minimize computational complexity, DAGC limits potential connections to an axial structure. For each patch, the algorithm calculates the Euclidean distance to other patches along its axis using a "roll" operation. A connection is established only if the distance is less than the difference between mean and standard deviation. DAGC allows for a variable number of connections per image. This means complex images can have more connections while simpler ones have fewer, enhancing representational power. This architecture is faster, more task specific and allows variable number of connections making it very practical. But it uses estimation rather than actual statistics which reduces trustness of the model. The model achieves 81.1\% top-1 accuracy, surpassing Vision GNN by 2.9\% and Vision HyperGraph Neural Network by 2.2\% on ImageNet classification. GreedyViG-B achieves 46.3 AP$_{\mathrm{box}}$ and 42.1 AP$_{\mathrm{mask}}$ on the COCO dataset.  On the ADE20K dataset, GreedyViG-B achieves 47.4 mIoU.

Tian et al. (2024) introduced Image Processing GNN (IPG)\cite{r26}, a novel model designed to advance image super-resolution by overcoming the rigidity of traditional methods. Standard techniques treat all pixels equally. However, IPG utilizes flexible graphs to account for the unbalanced nature of image reconstruction.  It uses a detail-rich indicator, that is DF, which measures the importance of a node by calculating the absolute difference between a feature map and its downsampled-then-upsampled version. IPG treats images as sets of pixel nodes rather than patch nodes. This finer-grained approach allows for direct pixel-to-pixel communication, which avoids the noise and errors. IPG uses two sampling strategies that gather both local and global information. This allows the model to rebuild lossy parts from their immediate neighborhood while also learning from distant but similar features across the whole image. The IPG model avoids misalignment issues, efficiently gathers both local and global information and DF is operationally cheap. But it has low inference speed as well as graph aggregation which means a potential information loss. IPG uses pixels as nodes instead of patches, the number of nodes is significantly larger. This creates a massive search space that makes it  impractical.Trained on DF2K\cite{r27} the model achieves state-of-the-art performance across various benchmarks, notably reaching a PSNR of 28.13dB on the Urban100\cite{r28} ×4 task.

Zhang et al. (2023) created a hybrid system called AF-GCN\cite{r29}  that combines Graph Convolution Network and Self-Attention mechanism. Self-Attention is great at finding long-range relationships but often miss small details and they are expensive to run. Graph Convolution Networks are great at finding small details and are faster, but they struggle with long-range dependencies. This experiment tries to capture both the tiny details and the big picture. The process begins with graph construction via ball queries, followed by a two-phase encoder: shallow stages use GCN blocks to capture local geometric details like edges and surfaces, while deeper stages employ Graph Attentive Filter (GAF) blocks to model long-range contexts by suppressing irrelevant noise from distant neighbors. To better handle spatial variations, a Spatial Feature Projection (SFP) module projects relative coordinates into the feature space within these convolutions. For data scaling, the model uses a graph-shared strategy where the graph structure used during Furthest Point Sampling (FPS) down-sampling is reused for graph de-convolution during up-sampling. Finally, because some points may lack information after up-sampling, trilinear interpolation and skip connections are used to restore fine details before an MLP-based classifier produces the final segmentation output. By using GCN blocks in early stages and Graph Attentive Filter (GAF) blocks in deeper stages, it avoids the high computational costs and local detail loss typical of pure attention-based networks. The graph-shared down-sampling and up-sampling strategy ensures that multi-scale graph structures are fully utilized throughout the network. The effectiveness of the model is highly dependent on how the feature learning phases are divided and it requires different configurations depending on the task.The model achieved state-of-the-art result on S3DIS\cite{r30}, highly competitive result on ScanNetV2\cite{r31}, Toronto-3D\cite{r32} and new performance record for object part segmentation on ShapeNetPart\cite{r33}.

\section{State-Space and Structured Approach}
\label{sec:approach4}
A State-Space Model is defined as a mathematical framework that describes a system with a set of state variables, which covers all the information needed to explain the systems functioning over time. It takes in equations that govern the system and relate observed outputs to the state. It originates from control theory. However, State Space Sequence Models\cite{r34} are a class of deep learning models designed for efficient sequence modeling, particularly for long-range dependencies. In order to make these models faster for deep learning training, structured matrices were introduced and it was called Structured State Space Sequence Models (S4)\cite{r35}. Orginally, State Space Models (SSMs) were designed for 1D sequential data. As computer vision data is 2D spatial data, researchers redesigned how images are processed with SSMs with the success of S4 and Mamba\cite{r36}. 

\subsection{State-of-the-Art State-Space and Structured Methods}
Traditional SSMs model sequential signals where a hidden state evolves over time according to the transition equation of states. In computer vision tasks, images are converted into sequences. Each patch embedding becomes an input to the state-space system. The hidden state updates using both the current input and previous state. This way the model can capture long-range spatial dependencies across an image. However, traditional SSMs face limitations in vision applications because flattening images into sequences disrupts inherent two-dimensional spatial relationships. It introduces directional bias.

Standard video transformers often struggle with the high computational costs and redundant data found in lengthy clips. To solve this problem, Wang et al. (2023) introduced a novel Selective Structured State-Space (S5) model\cite{r37} designed to improve the efficiency and accuracy of long-form video understanding. This begins by converting video frames into a sequence of image tokens using a video encoder, which are then processed by the proposed S5 model. Within this model, a momentum-updated S4 model generates global sequence-context features that guide a lightweight mask generator to calculate the importance of each token. This generator uses Gumbel-Softmax sampling to adaptively select only the most informative tokens, effectively discarding redundant data without the high computational cost of traditional dense self-attention. To further enhance the system, the authors introduce Long-Short Masked Contrastive Learning (LSMCL) as a pre-training step, where long and short video clips are randomly masked to mimic potential token-selection errors. This task specific selection reduces the memory footprint by around 25\%. There is also a risk of dropping informative tokens. The proposed S5 model achieves new state-of-the-art performance across three challenging benchmarks that are LVU\cite{r38}, COIN\cite{r39}, and Breakfast\cite{r40}, outperforming the previous state-of-the-art S4 model (ViS4mer) by up to 9.6\% in accuracy.

The Vision Mamba (Vim)\cite{r41} methodology addresses the challenges of position-sensitivity and global context requirements in visual understanding by proposing a pure State Space Model (SSM) backbone that processes image patches as a sequence with linear memory complexity and subquadratic computation time. The process begins by partitioning input images into flattened 2D patches, projecting them into vectors, and incorporating position embeddings to provide the spatial awareness necessary for location-aware visual recognition. At the core of the architecture, the Vim encoder utilizes bidirectional SSMs within each block to process the token sequence in both forward and backward directions, enabling data-dependent global visual context modeling without relying on self-attention. To maximize practical performance, the architecture integrates hardware-aware optimizations that reduce I/O overhead by managing data movement between high-bandwidth memory and fast SRAM, while also employing a recomputation strategy for intermediate activations to minimize the GPU memory footprint. Vim is better at capturing very large objects in images compared to window-based Transformers. By using bidirectional SSMs,  Vim captures data-dependent global visual context and directly learn sequential visual representations for dense prediction tasks while providing subquadratic-time computation and linear memory complexity. Finding the correct strategy for the bidirectional method is still difficult. Vim consistently outperforms well-established Vision Transformers like DeiT, achieving higher accuracy (e.g., +0.5 points on ImageNet and +0.9 mIoU on ADE20K), being 2.8× faster and saving 86.8\% GPU memory.

Liu et al. (2024) introduced VMamba\cite{r42} that begins with a stem module that partitions an input image into small 2D patches to form an initial feature map, which then undergoes hierarchical processing through multiple network stages involving down-sampling layers and Visual State-Space (VSS) blocks. At the heart of these VSS blocks is the 2D Selective Scan (SS2D) module, which overcomes the sequential limitations of standard 1D State Space Models by first performing a cross-scan to unfold the 2D feature map into four separate sequence paths traversing different directions. Each of these paths is processed independently and in parallel by its own S6 block using an input-dependent selection mechanism to capture contextual information with linear time complexity, after which a cross-merge operation reshapes and integrates these sequences back into a unified 2D map. This systematic approach establishes a global receptive field where every pixel can interact with every other pixel across various perspectives, and the overall framework is further refined across Tiny, Small, and Base scales through architectural optimizations. This model has linear time complexity, global receptive field and high inference throughput, making it suitable for practical applications although the models speed relies heavily on implementation enhancements. VMamba-Base attained a top-1 accuracy of 83.9\% on ImageNet-1K, achieved between 47.3\% and 49.2\% mAP on MSCOCO and  reached 51.0\% mIoU on ADE20K.

Ma. et al. (2024) addresses the critical challenge of effectively and efficiently modeling long-range dependencies in biomedical image segmentation. They introduced U-Mamba\cite{r43}, a hybrid CNN-SSM architecture. U-Mamba utilizes a hybrid encoder-decoder framework designed to efficiently integrate local feature extraction with long-range dependency modeling. The core building unit is the U-Mamba block, which consists of two successive Residual blocks to capture localized fine-grained features, followed by an SSM-based Mamba block. Within the Mamba block, multi-dimensional image features are flattened into a sequence and processed through two parallel branches: the first incorporates a linear expansion, a 1D convolution, and a selective SSM layer, while the second acts as a gating branch, with both eventually merging via a Hadamard product. It leverages a self-configuring mechanism inherited from the nnU-Net\cite{r44} framework, allowing the network to automatically adapt its configuration. To evaluate the performance Abdominal CT \& MRI\cite{r45}, Endoscopy images\cite{r46} and Microscopy images\cite{r47} were used. Experimental results demonstrate that U-Mamba consistently outperforms state-of-the-art CNN-based (nnU-Net, SegResNet) and Transformer-based (UNETR, SwinUNETR) networks.  The architecture proved to be more robust to heterogeneous appearances, achieving higher Dice Similarity Coefficients and F1 scores while producing significantly fewer segmentation outliers than its competitors.

Ruan et al. (2024) introduced VM-Unet\cite{r48}, which overcomes the computational limitations of Transformers by replacing the self-attention mechanism with State Space Models (SSMs) while maintaining long-range modeling capabilities. First, the model takes an image and breaks it into small, manageable squares called patches. In the Encoder phase, these patches pass through several stages where Visual State Space (VSS) blocks act like high-tech scanners. Because images are 2D but the Mamba "engine" processes sequences, the model uses a unique 2D-Selective-Scan (SS2D) technique: it scans the image from four different directions (like reading a page from top-left, bottom-right, top-right, and bottom-left simultaneously) to make sure it doesn't miss any global context. Inside these scans, a "selective mechanism" called the S6 block acts like a filter, automatically deciding which details are important for a diagnosis and which are just background noise. Patch Merging operations shrink the image size to help the model see the "big picture". To ensure fine details aren't lost, Skip Connections act as direct bridges, passing original sharp features from the early stages straight to the end of the process. Finally, the Decoder phase uses Patch Expanding to rebuild the image to its original size, layer by layer, while more VSS blocks refine the edges. Although efficient, the model requires further compression to be fully practical. VM-UNet demonstrates highly competitive performance in medical image segmentation, achieving DSC scores of 89.03 on ISIC17\cite{r49}, 89.71 on ISIC18\cite{r50,r51}, and 81.08 on the Synapse\cite{r52} dataset. In skin lesion tasks, it outperformed the powerful TransFuse hybrid model, achieving a 0.72\% increase in mIoU and a 0.44\% increase in DSC on the ISIC18 dataset while operating with significantly greater efficiency requiring only 4.11G FLOPs compared to the 11.50G used by TransFuse. 

\begin{table*}[t]
\centering
\small
\caption{Summary of State Space Model (SSM) based Vision Architectures}
\label{tab:ssm_vision_comparison}
\begin{tabularx}{\textwidth}{l p{1.8cm} p{2.2cm} X X X}
\toprule
\textbf{Paper} & \textbf{Model} & \textbf{Vision Task} & \textbf{Contribution} & \textbf{Strength} & \textbf{Limitation} \\ \midrule

Wang et al. (2023) \cite{r37} & S5 & Long-form video understanding & Updated S4 model specifically for long sequences & High efficiency in handling long-range dependencies & Risk of token loss during processing \\ \addlinespace

Zhu et al. (2024) \cite{r41} & Vim & Classification, Detection, Segmentation & Introduction of a pure SSM backbone for vision & Low memory usage compared to Transformers & Complex optimization requirements \\ \addlinespace

Liu et al. (2024) \cite{r42} & VMamba & Classification, Detection, Segmentation & Introduces the 2D Selective Scan (SS2D) mechanism & Global receptive field with linear complexity & Performance is highly implementation-dependent \\ \addlinespace

Ma et al. (2024) \cite{r43} & U-Mamba & Segmentation & Combines Residual blocks with SSM blocks for medical imaging & Highly robust across different segmentation tasks & Memory intensive during training phase \\ \addlinespace

Ruan et al. (2024) \cite{r48} & VM-UNet & Segmentation & Visual State Space blocks to scan images from four directions & Extremely efficient inference and processing & Needs further model compression for edge deployment \\ \addlinespace

Shi et al. (2025) \cite{r53} & VSSD & Classification, Detection, Segmentation & Removes token order dependency in SSMs & Linear global complexity with high throughput & Weakness in capturing fine-grained local details \\ 
\bottomrule
\end{tabularx}
\end{table*}

State Space Models (SSMs) and State Space Duality (SSD) have linear computational complexity; however, these models are inherently causal, which restricts their effectiveness in non-causal vision tasks. To address these limitations, Shi et al. (2025) propose the Visual State Space Duality (VSSD) model\cite{r53}.  The Visual State Space Duality (VSSD) methodology transforms the traditionally causal State Space Duality model into a non-causal format suitable for image processing. Instead of using the scalar parameter A to determine how much of a previous hidden state to keep, the researchers use it to dictate the specific contribution each individual token makes to a shared global hidden state. This shift removes the dependency on token order, effectively eliminating the "causal mask" and the need for complex multi-scan routes that were previously required to let tokens "see" the entire image. To maintain spatial awareness, the model incorporates 2D Rotary Position Embeddings (RoPE) and utilizes a hierarchical architecture that integrates these non-causal blocks with standard self-attention in the final stages to enhance high-level feature processing. VSSD maintains linear complexity, has a global receptive field without quadratic cost and preserves spatial relationships through positional encoding. But the model can struggle with fine-grained local details, has high-level processing and relies on external techniques to introduce beneficial inductive biases. VSSD-T achieved a top-1 accuracy of 83.8\%, outperforming VMamba-T by 1.2\%, outperformed the Swin-T transformer by margins of +4.3 in box AP and +3.5 in mask AP, achieved 48.7 mIoU (single-scale), surpassing Swin-T, ConvNeXt-T, and VMamba-T by 4.3, 2.7, and 0.7 points, respectively.

\section{Implicit Neural Representation Models}
\label{sec:approach5}
Implicit Neural Representations (INRs) are a paradigm where data is represented as a continuous function parameterized by a neural network, mapping coordinates to signals. This idea gained significant traction in 3D vision after Mildenhall et al. (2020) introduced Neural Radiance Fields (NeRF)\cite{r54}. It needed no explicit geometry, no mesh, just a network that learns what the scene looks like. Since then, researchers have taken INRs in many different directions, tackling persistent issues around surface quality, rendering speed, hardware compatibility, and robustness under challenging capture conditions.

\subsection{State-of-the-Art INR Methods}
One of the earliest and most persistent problems in this space is recovering fine geometric detail from only RGB images, without relying on depth sensors. Li et al. (2023) addressed this with Neuralangelo\cite{r55}. It represents the scene using a Signed Distance Function (SDF) paired with multi-resolution hash encoding\cite{r56}. Hash stores positional data at multiple resolutions. Neuralangelo computes gradients by sampling a few nearby points and approximates the gradient from those. The surfaces that come out of this are noticeably more stable, analytical gradients have a tendency to produce noise around sharp edges, which this approach largely avoids. Training starts coarse and gets progressively more detailed, which stops the model from overfitting to noise before the overall geometry is sorted out. On the DTU benchmark\cite{r57}, the method logged a Chamfer distance of 0.61 mm and PSNR of 33.84, coming out ahead of both NeuS\cite{r58} and NeuralWarp. Results on the Tanks and Temples dataset\cite{r59} further confirmed its advantage in large-scale outdoor scenes. The downside is computational cost, the combination of numerical gradients and multiple loss terms makes training slow, and the method is far from suitable for any interactive application.

\begin{table*}[t]
\centering
\small
\caption{Summary of Implicit Neural Representation Models}
\label{tab:3d_reconstruction_comparison}
\begin{tabularx}{\textwidth}{l p{2cm} p{2.2cm} X X X}
\toprule
\textbf{Author} & \textbf{Model} & \textbf{Vision Task} & \textbf{Contribution} & \textbf{Strength} & \textbf{Limitation} \\ \midrule

Li et al. (2023) \cite{r55} & Neuralangelo & 3D Surface Reconstruction & SDF + multi-resolution hash encoding with numerical gradients & Stable, detailed surfaces; progressive coarse-to-fine training & Slow training; not suitable for interactive applications \\ \addlinespace

Barron et al. (2023) \cite{r60} & Zip-NeRF & Image-Based 3D Scene Reconstruction & 3D volume-based sampling with Gaussian-weighted sub-samples & Reduces aliasing; 24$\times$ faster training than mip-NeRF 360 & $\sim$6$\times$ slower than iNGP; requires high-end GPU \\ \addlinespace

Kerbl et al. (2023) \cite{r62} & 3D Gaussian Splatting & 3D Rendering & Scene represented as 3D Gaussians with tile-based GPU rasterizer & First real-time high-quality rendering at 1080p (30+ FPS) & High memory usage ($>$20 GB); artifacts in sparse regions \\ \addlinespace

Low and Lee (2024) \cite{r63} & Deblur e-NeRF & 3D Scene Reconstruction & Physically grounded pixel bandwidth model for event camera blur & Strong performance under high-speed and low-light conditions & Requires known camera trajectories for optimal results \\ \addlinespace

Chen et al. (2023) \cite{r64} & MobileNeRF & Novel View Synthesis & Converts NeRF to textured polygons for standard GPU rasterization & Runs on mobile (55 FPS on iPhone XS) & Struggles with specular/transparent surfaces; fixed mesh resolution \\ 
\bottomrule
\end{tabularx}
\end{table*}

Surface reconstruction and novel view synthesis are closely related but distinct goals. The latter requires not only correct geometry but also accurate appearance across different scales and viewing distances, which is where many fast NeRF variants struggle. Barron et al. (2023) built Zip-NeRF\cite{r60} around a fairly specific frustration with grid-based methods like Instant NGP, if you only query one point per ray, you end up blind to structure at other scales, and the result shows up as aliasing whenever the camera shifts or zooms. Their fix was to stop treating samples as points altogether. Each sample is instead modeled as a small 3D volume, broken down into a group of Gaussian-weighted sub-samples that get averaged before being fed into the network. A six-point multisampling pattern along each ray ensures the volume is properly covered, and a new interlevel loss targets aliasing along the depth axis specifically,  something that prior methods had mostly overlooked. The gains over mip-NeRF 360\cite{r61} are meaningful: DSSIM drops by 17\% and LPIPS by 19\%, while training is roughly 24 times faster. Against iNGP the improvements are even larger. The remaining limitation is that Zip-NeRF is still around 6× slower than iNGP and demands high-end GPU hardware for training.

Real-time rendering had long been treated as fundamentally incompatible with high-quality neural scene representations. Kerbl et al. (2023) changed that perception with 3D Gaussian Splatting\cite{r62}, which avoids ray marching altogether by representing the scene as millions of 3D Gaussians. Each Gaussian is pinned to a location in 3D space and comes with four things attached to it: an opacity value, a covariance matrix that stretches or squashes it into different shapes, and a set of spherical harmonic coefficients that encode how its color shifts depending on the viewing angle. Rendering is done by projecting these Gaussians onto the image plane and compositing them using a tile-based GPU rasterizer that sorts and blends them front-to-back, which also supports gradient flow for optimization. To keep the representation efficient, an adaptive densification strategy runs throughout training, small Gaussians in under-reconstructed regions are cloned, oversized ones are split, and nearly transparent Gaussians are pruned. Training wraps up in 35–45 minutes compared to roughly 48 hours for Mip-NeRF360, and the final model renders at over 30 FPS at 1080p resolution. This was the first time real-time, high-quality novel view synthesis was demonstrated at that resolution. High memory usage, often exceeding 20 GB, and occasional artifacts in sparsely observed regions are the main concerns that remain open.

Most methods in this family take for granted that input images are sharp and well-exposed, which is a reasonable assumption in controlled settings but breaks down quickly in practice. Low and Lee (2024) designed Deblur e-NeRF\cite{r63} for  reconstructing NeRFs from event camera data captured under high-speed motion or very low light. Event cameras work by recording per-pixel brightness changes asynchronously. However, the limited bandwidth of each pixel's analog circuit introduces its own form of motion blur that standard NeRF training simply ignores. The central contribution of Deblur e-NeRF is a physically grounded pixel bandwidth model that describes how each pixel's photoreceptor, buffer, and amplifier stages respond to incoming light over time. This model is linearized and discretized to make training tractable, and a threshold-normalized total variation loss is added to stabilize reconstruction across large textureless regions. A translated-gamma correction is applied after reconstruction to account for unknown black levels and potential gamma inaccuracies. Even under the hardest combined conditions of high speed and low light, Deblur e-NeRF achieved PSNR 25.59 and SSIM 0.896, substantially ahead of competing event-based baselines. The method requires known camera trajectories, which somewhat limits deployment in fully unconstrained settings.

Even setting aside quality and speed, a more practical barrier to adoption is that most neural rendering pipelines are simply too heavy to run on consumer hardware. Chen et al. (2023) approached this from a systems perspective with MobileNeRF\cite{r64}, which converts the trained neural representation into textured polygons that slot directly into standard GPU rasterization pipelines. The conversion happens over three training stages: a continuous NeRF-like model is trained first, opacities are then binarized to match hardware-friendly rasterization, and finally the mesh is discretized with features and opacities baked into PNG texture maps. At render time, the scene is drawn using z-buffering, exactly the same pipeline that powers real-time game engines, and a small MLP in the fragment shader maps per-pixel features and view directions to final colors. Super-sampling is applied at the feature level rather than the color output, cutting computation while retaining detail. On a real mobile device, an iPhone XS rendered synthetic 360° scenes at 55 FPS with only 538 MB of GPU memory, while SNeRG\cite{r65} frequently failed to run at all due to memory constraints. Specular and semi-transparent surfaces remain problematic because of the binary opacity assumption, and fixed mesh resolution can cause detail to break down on close-up views, but for a broad class of real-world scenes MobileNeRF offers a genuinely practical path to deployment.

\section{Energy-Based and Kernel-Driven Models}
\label{sec:approach6}
Kernel-driven methods swap out fixed convolution filters for ones that reshape themselves. Both paradigms are harder to train and slower at inference but they handle ambiguity and geometric variation which is beneficial.

\subsection{State-of-the-Art Energy-Based and Kernel-Driven Methods}
Chen et al. (2023) built DiffusionDet\cite{r66} with the idea to treat detection as a denoising problem. A random noise is added in training so the model learns how to remove it. A backbone encoder extracts features, and a multi-stage detection decoder refines a set of noisy boxes. Boxes under threshold are replaced with freshly sampled random boxes through a renewal step. The model self corrects in this manner. On COCO, DiffusionDet with ResNet-50 scored 45.8 AP, climbing to 46.8 with more boxes and steps, and hitting 53.3 AP with a Swin-Base backbone, clearing both Faster R-CNN and DETR. On LVIS, it scored 29.4 AP with ResNet-50, improving to 33.0 AP with more boxes and iterations, and reaching 42.0 AP with Swin-Base. It also transferred to CrowdHuman\cite{r67} without any retraining, eventually reaching 91.4 AP50 after fine-tuning. The model runs at 30 FPS, though iterative refinement still adds latency that one-shot detectors simply do not have.

\begin{table*}[t]
\centering
\small
\caption{Summary of Energy-Based and Kernel-Driven Models}
\label{tab:diffusion_vision_comparison}
\begin{tabularx}{\textwidth}{l p{1.8cm} p{2.2cm} X X X}
\toprule
\textbf{Paper} & \textbf{Model} & \textbf{Vision Task} & \textbf{Contribution} & \textbf{Strength} & \textbf{Limitation} \\ \midrule

Chen et al. (2023) \cite{r66} & DiffusionDet & Object Detection & Detects objects by denoising random boxes step by step; bad boxes replaced with new ones & High AP on COCO, LVIS, CrowdHuman; flexible box count; 30 FPS & Slower than one-shot detectors; lacks deformable attention \\ \addlinespace

Xie et al. (2024) \cite{r68} & DiffusionTrack & Visual Object Tracking & Tracks target as a point set refined via global interaction and confidence voting & Top results on GOT-10k, LaSOT; early exit gives 16.7\% faster inference & Hard cases need deeper decoder, which slows down speed \\ \addlinespace

Yi et al. (2023) \cite{r72} & Diff-Retinex & Image Enhancement & Splits image into reflectance/illumination; restores each using diffusion & Best FID, LPIPS on LOL datasets; recovers texture and color well & PSNR lower than LLFormer; expensive training; sensitive hyperparameters \\ \addlinespace

Zbinden et al. (2023) \cite{r74} & CCDM & Image Segmentation & Learns distribution of segmentation outputs using categorical noise and U-Net & State-of-the-art on LIDC with only 9M parameters; handles uncertainty & Slow inference due to iterative sampling; hard to scale to high-res \\ \addlinespace

Zheng et al. (2024) \cite{r77} & KACM & Scene Text Detection & Uses multiple kernels per location and predicts distance maps for full regions & Strong F-measure on Total-Text; handles curved and overlapping text & No gain beyond 4 kernels; struggles with very small or dense text \\ 
\bottomrule
\end{tabularx}
\end{table*}
As the iterative refinement idea applies equally well to temporal problems, Xie et al. (2024) extended it to visual object tracking with DiffusionTrack\cite{r68}. One-shot trackers commit to a location immediately, which breaks down when the scene has distractors or the target changes appearance. DiffusionTrack represents the target as a set of points spread across the object and refining points. A Vision Transformer encoder processes the template and search frame. The diffusion decoder then runs through three operations per step: global interaction between all points, dynamic convolution to pull in spatial features from the encoder, and a localization refinement that predicts updated positions and confidence scores. Low-confidence points are swapped out for new random samples while high-confidence ones are voted on to filter out distractors. On GOT-10k\cite{r69}, DiffusionTrack reached AO 75.2\% and SR0.75 72.0\%, ahead of previous state-of-the-art trackers. On LaSOT\cite{r70} it scored AUC 72.3\%, on TrackingNet\cite{r71} it ranked first with AUC 85.2\%, and on TNL2K it achieved AUC 56.8\%. An early exit strategy enables 16.7\% faster inference with minimal accuracy loss. Getting the speed up without sacrificing accuracy on difficult sequences remains an open problem, deeper decoder configurations help on hard cases but push the frame rate down noticeably.

Diffusion models turn out to be just as useful in low-level vision, where the goal is not to locate objects but to recover degraded or missing image content. Low-light enhancement is trickier than it looks, it is not just about brightening an image, but recovering detail that the sensor never properly captured in the first place. Yi et al. (2023) approached this with Diff-Retinex\cite{r72}, which borrows the Retinex idea of splitting a scene into what it looks like and how it is lit, then uses a diffusion model to fix both parts separately. A Transformer network handles the decomposition, separating the input into a reflectance component and an illumination component. Each one then goes through its own diffusion process, noise is added and gradually removed, and in that process the model fills gaps, corrects color, and reduces noise. The two cleaned-up maps are multiplied back together to produce the final output. On the LOL dataset\cite{r73}, Diff-Retinex achieved FID 47.85, LPIPS 0.048, and BIQI 19.97, outperforming RetinexNet, KinD, and URetinex across all perceptual metrics. On VE-LOL-L, FID reached 47.75, LPIPS 0.050, and BIQI 26.54, showing strong generalization across datasets. PSNR came in at 21.98, marginally below LLFormer's 23.66, but competitive with or better than all other Retinex-based methods. Training is expensive and the model is sensitive to hyperparameter choices, which makes it less straightforward to deploy outside a controlled research environment.

Standard diffusion models operate over continuous values, which raises the question of whether the same idea can work for categorical outputs like segmentation labels. Zbinden et al. (2023) showed that it can, through CCDM\cite{r74}, a Conditional Categorical Diffusion Model built for stochastic semantic segmentation. CCDM learns the full distribution over possible segmentation maps rather than a single prediction. The forward process corrupts clean label maps with categorical noise, and a U-Net with self-attention learns the reverse, conditioned on either raw image pixels or DINOv2 pretrained features. On the LIDC lung nodule dataset\cite{r75}, CCDM achieved GED16 of 0.212 and HM-IoU16 of 0.623 using only 9M parameters, outperforming 11 stochastic segmentation baselines including MoSE which requires 42M parameters. On LIDCv2, it achieved GED16 of 0.239 and HM-IoU16 of 0.598. On Cityscapes\cite{r76}, CCDM-Dino reached mIoU 65.8 at 256×512 resolution. The main limitation is computational, iterative sampling is expensive, and scaling to high-resolution images without a smarter categorical compression strategy remains a challenge.
\begin{table*}[b]
\centering
\small
\caption{Summary of Uncertainty-Aware Vision Models}
\label{tab:uncertainty_vision_comparison}
\begin{tabularx}{\textwidth}{l p{2cm} p{2.2cm} X X X}
\toprule
\textbf{Paper} & \textbf{Model} & \textbf{Vision Task} & \textbf{Contribution} & \textbf{Strength} & \textbf{Limitation} \\ \midrule

Franchi et al. (2024) \cite{r84} & ABNN & Classification and Segmentation & Principled uncertainty estimation framework for neural networks & Scalable, reliable, and practical for real-world deployment & High dependency on the quality of pretraining \\ \addlinespace

Kai et al. (2023) \cite{r86} & U2MOT & Object Tracking & Integration of uncertainty estimation into the Multi-Object Tracking (MOT) pipeline & Reliable and highly compatible with existing tracking frameworks & Heavy dependency on the accuracy of initial object detections \\ 
\bottomrule
\end{tabularx}
\end{table*}
Zheng et al. (2024) developed the Kernel Adaptive Convolution Module (KACM)\cite{r77}. Texts may vary. It may have different size, background, style and a fixed-kernel convolution may not respond well. KACM picks from several kernels of different sizes at every spatial location blending their outputs, so the features it produces naturally scale with the text. It predicts a distance map alongside the usual segmentation output. It records how close each pixel sits to the nearest text edge, which gives the decoder a much more detailed picture of where the boundaries actually fall. At inference, text centers are extracted by thresholding this distance map, and a Vatti clipping algorithm expands them outward to cover full text regions. On Total-Text\cite{r78}, the model reached an F-measure of 90.98\% at 25.0 FPS, and on TD500\cite{r79}, 90.79\% F-measure at 18.1 FPS, with strong results also reported on CTW1500 and ICDAR19 ArT. Adding more than four kernels did not improve results meaningfully, and very small or heavily crowded text regions still present difficulties that the current design does not fully address.

\section{Probabilistic \& Uncertainty-Awareness }
\label{sec:approach7}
CV models that think about probability and uncertainty show how sure they are about predictions. It is important in safety-critical domains. Uncertainty caused by limited knowledge of model parameters is epistemic uncertainty\cite{r80} and noise in the input data is aleatoric uncertainty\cite{r81}. 

\subsection{State-of-the-Art Probabilistic and Uncertainty-Aware Methods}
Standard approaches for deep learning uncertainty include ensemble methods, MC dropout, BNNs, or augmenting deterministic networks with confidence heuristics \cite{r81,r82}. But they introduce complexity and need heavy computation. BNNs estimate uncertainty by placing probability distributions over model weights, which offers a theoretically principled framework but is difficult to scale to large modern architectures due to expensive inference and training procedures\cite{r83}.

Adaptable Bayesian Neural Network (ABNN) framework, through a post-hoc adaptation, converts a pre-trained deterministic neural network into a Bayesian model\cite{r84}. It inserts stochastic Bayesian layers around normalization layers in a pretrained architecture and fine-tunes only these additional components. During inference, predictive uncertainty is estimated through multiple stochastic forward passes. This design preserves the computational efficiency of the original model while enabling probabilistic inference. Empirical results demonstrate that ABNN maintains competitive predictive performance while significantly improving uncertainty calibration and out-of-distribution detection. For example, the approach achieves approximately 76\% Top-1 accuracy on ImageNet\cite{r7}, 97\% accuracy on CIFAR-10\cite{r16}, and 81\% accuracy on CIFAR-100\cite{r85}, while outperforming common uncertainty estimation baselines such as MC Dropout and last-layer Bayesian models. Despite these advantages, ABNN still relies on approximate Bayesian inference and requires multiple inference passes to estimate uncertainty, which slightly increases computational overhead.

Another recent line of work addresses uncertainty in unsupervised multi-object tracking (MOT). Conventional unsupervised tracking methods typically rely on pseudo tracklets derived from detection associations across frames, assuming these associations to be correct during training. However, such pseudo-labels are frequently unreliable due to occlusions, detection errors, or abrupt appearance changes. To mitigate this problem, the U2MOT framework introduces explicit uncertainty modeling into the pseudo-label generation process\cite{r86}. The method first estimates association uncertainty to measure the reliability of tracklet matches between frames. Based on this measure, it performs uncertainty-aware tracklet labeling, which filters out ambiguous or unreliable pseudo-labels while retaining high-confidence associations for training.

In addition, U2MOT introduces tracklet-guided augmentation, which generates realistic motion variations from reliable tracklets, and a hierarchical sampling strategy that prioritizes trustworthy samples while still exposing the model to difficult cases. This uncertainty-aware training pipeline produces cleaner supervision signals and improves identity consistency across frames. Experimental evaluations on the MOT17\cite{r87}, MOT20\cite{r88}, and VisDrone-MOT\cite{r89} benchmarks demonstrate that U2MOT achieves state-of-the-art performance among unsupervised tracking approaches, reaching 64.2\% HOTA on MOT17 and 62.7\% HOTA on MOT20, surpassing earlier methods such as SimUMOT and UEANet . Nevertheless, the approach remains dependent on the quality of the initial object detections and may degrade under severe occlusion or poor detection conditions.

\section{Classical Computer Vision–Deep Learning Hybrid Models}
\label{sec:approach8}
Classical computer vision-deep learning hybrid models try to gather the advantages of traditional computer vision techniques and modern deep learning architectures. Classical vision methods are built on geometric constraints, correspondence matching, robust estimation, rendering priors, and structured pipelines. Deep learning methods excel at learning complex representations directly from big data. Hybrid models integrate learning-based components into classical pipelines in order to improve performance while preserving the interpretability and structural advantages of traditional vision algorithms.

\subsection{State-of-the-Art Classical CV-Deep Hybrid Methods}
Historically, many computer vision pipelines were composed of carefully engineered modular components. Feature matching relied on nearest-neighbour descriptor comparisons followed by heuristic filtering; robust geometric estimation was performed using the RANSAC algorithm. Image generation and segmentation depended on task-specific architectures and dynamic scene reconstruction relied on computationally expensive volumetric representations such as Neural Radiance Fields (NeRF) \cite{r90,r91}. These approaches provided strong geometric foundations. But they often struggle with practical application when confronted with complex real-world data. Integrating these pipelines with deep learning has brought about good results.

One prominent example is ControlNet, which introduces structured conditioning into diffusion-based image generation models\cite{r92}. Traditional text-to-image diffusion models, rooted in denoising diffusion probabilistic models (DDPM)\cite{r100} and later improved through latent diffusion models (LDM) \cite{r101}, generate images but may lack precise spatial control because textual descriptions alone cannot fully specify layout, pose, or scene geometry. ControlNet tries to address this limitation by attaching a conditional control branch to a pretrained diffusion backbone. The control branch is connected through zero-initialized convolutions that gradually learn to inject structural guidance signals such as edges, human poses, segmentation maps, or depth cues. This architecture enables controllable image generation while preserving the pretrained model’s capabilities. Experimental results show that ControlNet significantly improves spatial controllability and visual fidelity across datasets of varying sizes.

Another major development in generative modeling is InstaFlow, which tackles the slow inference speed of traditional diffusion models\cite{r93}. Conventional diffusion frameworks require dozens of iterative denoising steps to generate an image, which limits their applicability in real-time settings. InstaFlow reformulates diffusion generation as a rectified probability flow and applies a reflow process that simplifies the diffusion trajectory into a nearly linear mapping from noise to image. A student network is then trained to approximate this mapping in a single forward pass, effectively distilling a multi-step diffusion process into a one-step generator. InstaFlows reduced latency reached FID 23.3 in a single step on MS COCO 2017-5k\cite{r24}.

Feature matching is being progressed. Traditional pipelines rely on handcrafted descriptors and nearest-neighbor search. Deep matching methods such as SuperGlue\cite{r102} and LoFTR\cite{r103} improved robustness but introduced higher computational costs. LightGlue addresses this challenge by introducing an adaptive transformer-based matching architecture that balances accuracy and efficiency\cite{r94}. The model processes keypoints and descriptors from two images using self-attention and cross-attention layers, enabling global reasoning about correspondences. Furthermore, it employs adaptive depth and adaptive width pruning, allowing the network to terminate early for easy image pairs and discard unlikely keypoints to reduce computational load. Experimental results demonstrate that LightGlue achieves state-of-the-art matching accuracy while operating approximately 2.5 times faster than SuperGlue.
\begin{table*}[t]
\centering
\small
\caption{Summary of Hybrid Vision Models}
\label{tab:gen_geom_vision_comparison}
\begin{tabularx}{\textwidth}{l p{1.8cm} p{2.2cm} X X X}
\toprule
\textbf{Paper} & \textbf{Model} & \textbf{Vision Task} & \textbf{Contribution} & \textbf{Strength} & \textbf{Limitation} \\ \midrule

Zhang et al. (2023) \cite{r92} & ControlNet & Image Generation & Adds structural conditioning to large-scale diffusion models & Fine-grained spatial control over generation & Highly dependent on the quality of pretraining \\ \addlinespace

Liu et al. (2023) \cite{r93} & InstaFlow & Image Generation & Distills multi-step diffusion into single-step generation & Extremely fast inference speed & Slight drop in visual quality compared to multi-step diffusion \\ \addlinespace

Lindenberger et al. (2023) \cite{r94} & LightGlue & Feature Matching & Introduces adaptive depth pruning for neural matching & Fast and accurate across diverse scenes & Significantly heavier than classical matching methods \\ \addlinespace

Brachmann \& Rother (2019) \cite{r96} & NG-RANSAC & Geometric Estimation & Makes RANSAC differentiable for end-to-end learning & Improves robustness to outliers & High training complexity; remains inherently stochastic \\ \addlinespace

Wang et al. (2023) \cite{r97} & YOLOv7 & Object Detection & Optimized trainable bag-of-freebies and architecture scaling & State-of-the-art real-time high accuracy & Less flexible than modern transformer-based detectors \\ \addlinespace

Wu et al. (2024) \cite{r98} & 4DGS & Neural Rendering & Extends Gaussian splatting to the temporal/4D domain & Very fast rendering for dynamic scenes & Complex deformation modeling requirements \\ \addlinespace

Kirillov et al. (2023) \cite{r99} & SAM & Segmentation & Foundation model for promptable image segmentation & Exceptional zero-shot generalization & Requires significant GPU memory for inference \\ \addlinespace

Xiao et al. (2024) \cite{r105} & SpatialTracker & Object Tracking & Tracks objects through 3D geometry consistency & Robust to occlusion and long-term motion & Performance depends heavily on depth estimation quality \\ 
\bottomrule
\end{tabularx}
\end{table*}
Another important direction involves integrating learning into robust geometric estimation. The classical RANSAC algorithm remains a cornerstone of many vision pipelines but relies on non-differentiable random sampling and hard hypothesis selection, preventing end-to-end learning\cite{r95}. NG-RANSAC (Neural-Guided RANSAC) addresses this limitation by introducing a differentiable formulation of the RANSAC pipeline\cite{r96}. The method relaxes the sampling procedure so that gradients can propagate through hypothesis generation and evaluation. A learned quality function aggregates information from multiple hypotheses rather than selecting a single best solution, enabling the model to learn better sampling strategies and inlier predictions. This approach improves geometric estimation accuracy while maintaining runtime comparable to classical RANSAC implementations.

In the domain of real-time object detection, YOLOv7 demonstrates that substantial performance improvements can be achieved through better training strategies rather than simply increasing model complexity\cite{r97}. This direction is complementary to transformer-based detectors such as DETR\cite{r104}, which reformulate detection as a set prediction problem using attention mechanisms inspired by vision transformers. YOLOv7 introduces Extended Efficient Layer Aggregation Networks (E-ELAN) to enhance gradient propagation and enable more effective scaling of network depth and width. Additional techniques such as planned re-parameterization and improved label assignment further strengthen training efficiency. YOLOv7 achieves 56.8\% average precision on the MS COCO with significantly lower computational cost.

Hybrid approaches have also transformed scene representation and neural rendering. 4D Gaussian Splatting (4D-GS) extends the efficiency of Gaussian splatting to dynamic scenes by learning a spatiotemporal deformation model\cite{r98}. Instead of representing each frame independently, the method maintains a canonical set of 3D Gaussians and learns a deformation field that transforms them over time. A spatial-temporal encoder captures motion patterns, while a lightweight neural decoder predicts time-specific Gaussian transformations. This design enables compact dynamic scene representation while maintaining the high rendering speed of rasterization-based splatting techniques. Experiments show that 4D-GS can achieve up to 82 frames per second at 800×800 resolution while maintaining competitive rendering quality.

Another influential hybrid model is the Segment Anything Model (SAM), which reframes image segmentation as a prompt-based task analogous to prompting in large language models\cite{r99}. The architecture consists of an image encoder, a prompt encoder that embeds user inputs such as points or bounding boxes, and a lightweight mask decoder that produces segmentation masks conditioned on these prompts. To support large-scale training, the authors constructed the SA-1B dataset, which contains over one billion segmentation masks from eleven million images. Trained on this dataset, SAM demonstrates strong zero-shot segmentation capabilities across multiple benchmarks without requiring task-specific retraining.

Finally, SpatialTracker extends classical tracking approaches by incorporating three-dimensional geometric reasoning into deep tracking models\cite{r105}. Traditional optical flow and feature tracking methods operate primarily in the 2D image plane, which makes them sensitive to occlusions and out-of-plane motion. Modern flow-based methods such as RAFT \cite{r106} have improved dense correspondence estimation, but still operate largely in 2D. SpatialTracker addresses this limitation by lifting pixel coordinates into 3D space using monocular depth estimation and representing scenes using a triplane feature structure.

\section{Limitation}
This review has several weaknesses associated with it. First of all, it relies on a selection of 40 articles chosen as an effort to include both classic and recent works; however, it is possible that it failed to incorporate other important studies related to computer vision. In addition, the very selection procedure may create a bias towards particular sources, databases, and key words used in searching the literature. Moreover, the rapid development of computer vision may result in omission of some recent technologies. Individual research papers were analyzed from the standpoint of methodology, strengths, and weaknesses; however, the review is a purely qualitative one without any meta-analysis or quantifying involved.

\section{Future Direction}
Several important research directions emerge from this taxonomy that warrant deeper investigation. First, the addition of state-space models with convolution or graph-based models present an opportunity for building architectures that are efficient, local-structure-aware, and capable of long-range reasoning. Second, implicit neural representations remain underexplored as a backbone for discriminative tasks. Third, probabilistic and uncertainty-aware architectures need to move towards reasoning for safety-critical domains. Fourth, the hybrid classical-deep category invites revisitation for data scarcity and low material availability conditions. Finally, a cross-category direction lies in developing unified benchmarks that evaluate not only accuracy but also computational efficiency, robustness to distribution shift, uncertainty calibration, and interpretability simultaneously, a criteria that would allow fairer and more comprehensive comparison across this heterogeneous landscape of transformer alternatives.

\section{Conclusion}
This survey outlines a structured taxonomy of eight categories of transformer-free architectures in computer vision, ranging from convolutional and MLP-based models to graph-based, state-space, and structured approaches, as well as implicit neural representations, energy-based and kernel-driven methods, and probabilistic and hybrid frameworks. Each alternative has its own advantages and they are suited for particular kinds of problems. Convolutional and MLP-based models continue to show how powerful spatial locality and weight sharing are. Graph-based and state-space models build on these ideas, extending them to more complex settings such as non-Euclidean data and long-range dependencies, often with better computational scaling. In a different direction altogether, implicit neural representations and energy-based approaches rethink how visual information is modeled, either as continuous functions or as distributions defined by energy landscapes. Probabilistic and hybrid methods bring attention to concerns that are sometimes overlooked, especially uncertainty estimation and reliability to use in the real world. This taxonomy suggests that architectural diversity is a legacy of earlier approaches as well as an essential and active force shaping computer vision today.

{
    \small
    \bibliographystyle{unsrtnat}
    \bibliography{main}
}


\end{document}